%% file: 0-main.tex
\documentclass[acmtog]{acmart} %
\setcopyright{acmlicensed}
\AtBeginDocument{%
  }

\acmSubmissionID{768}

\usepackage{graphicx}
\usepackage{amsmath}
\usepackage{booktabs}
\usepackage{lib}
\usepackage{mathtools}
\usepackage{float}

\input{macro}

\def \cameraready {}

\ifx \showeditI \undefined
    \newcommand{\editI}[1]{{#1}}
    \newcommand{\editII}[1]{{#1}}
\else
    \newcommand{\editI}[1]{{\color{blue}{#1}}}
    \newcommand{\editII}[1]{{\color{blue}{#1}}}
\fi

\ifx \showeditII \undefined
    \relax %
\else
    \let\editI\undefine
    \let\editII\undefine

    \newcommand{\editI}[1]{{#1}}
    \newcommand{\editII}[1]{{\color{blue}{#1}}}
\fi

\ifx \cameraready \undefined
    \relax %
\else
    \let\editI\undefine
    \let\editII\undefine

    \newcommand{\editI}[1]{{#1}}
    \newcommand{\editII}[1]{{#1}}
\fi

\ifx \submission \undefined

\else

\fi

\newcommand{\modelname}[1]{3D-Fixup}

\begin{document}

\title{3D-Fixup: Advancing Photo Editing with 3D Priors}

\author{Yen-Chi Cheng}
\authornote{Work was done while Yen-Chi was an intern at Adobe Research.}
\email{yenchic3@illinois.edu}
\affiliation{%
  \institution{University of Illinois Urbana-Champaign}
  \city{Urbana}
  \state{Illinois}
  \country{USA}
}

\author{Krishna Kumar Singh}
\email{krishsin@adobe.com}
\affiliation{%
  \institution{Adobe Research}
  \city{San Jose}
  \state{California}
  \country{USA}
}
\author{Jae Shin Yoon}
\email{jaeyoon@adobe.com}
\affiliation{%
  \institution{Adobe Research}
  \city{San Jose}
  \state{California}
  \country{USA}
}
\author{Alexander Schwing}
\email{aschwing@illinois.edu}
\affiliation{%
  \institution{University of Illinois Urbana-Champaign}
  \city{Urbana}
  \state{Illinois}
  \country{USA}
}
\author{Liang-Yan Gui}
\email{lgui@illinois.edu}
\affiliation{%
  \institution{University of Illinois Urbana-Champaign}
  \city{Urbana}
  \state{Illinois}
  \country{USA}
}
\author{Matheus Gadelha}
\email{gadelha@adobe.com}
\affiliation{%
  \institution{Adobe Research}
  \city{San Jose}
  \state{California}
  \country{USA}
}
\author{Paul Guerrero}
\email{guerrero@adobe.com}
\affiliation{%
  \institution{Adobe Research}
  \city{London}
  \country{UK}
}
\author{Nanxuan Zhao}
\email{nanxuanz@adobe.com}
\affiliation{%
  \institution{Adobe Research}
  \city{San Jose}
  \state{California}
  \country{USA}
}

\begin{abstract}
Despite significant advances in modeling image priors via diffusion models, 3D-aware image editing remains challenging, in part because the object is only specified via a single image.
To tackle this challenge, we propose \modelname{}, a new framework for editing 2D images guided by learned 3D priors. The framework supports difficult editing situations such as object translation and 3D rotation.
To achieve this, we leverage a training-based approach that harnesses the generative power of diffusion models. As video data naturally encodes real-world physical dynamics, we turn to video data for generating training data pairs, i.e., a source and a target frame. Rather than relying solely on a single trained model to infer transformations between source and target frames, we incorporate 3D guidance from an Image-to-3D model, which bridges this challenging task by explicitly projecting 2D information into 3D space. 
We design a data generation pipeline to ensure high-quality 3D guidance throughout training. Results show that by integrating these 3D priors, \modelname{} effectively supports complex, identity coherent 3D-aware edits, achieving high-quality results and advancing the application of diffusion models in realistic image manipulation. \editI{The code is provided at \href{https://3dfixup.github.io/}{https://3dfixup.github.io/}}.
\end{abstract}


\begin{teaserfigure}
  \includegraphics[width=\linewidth]{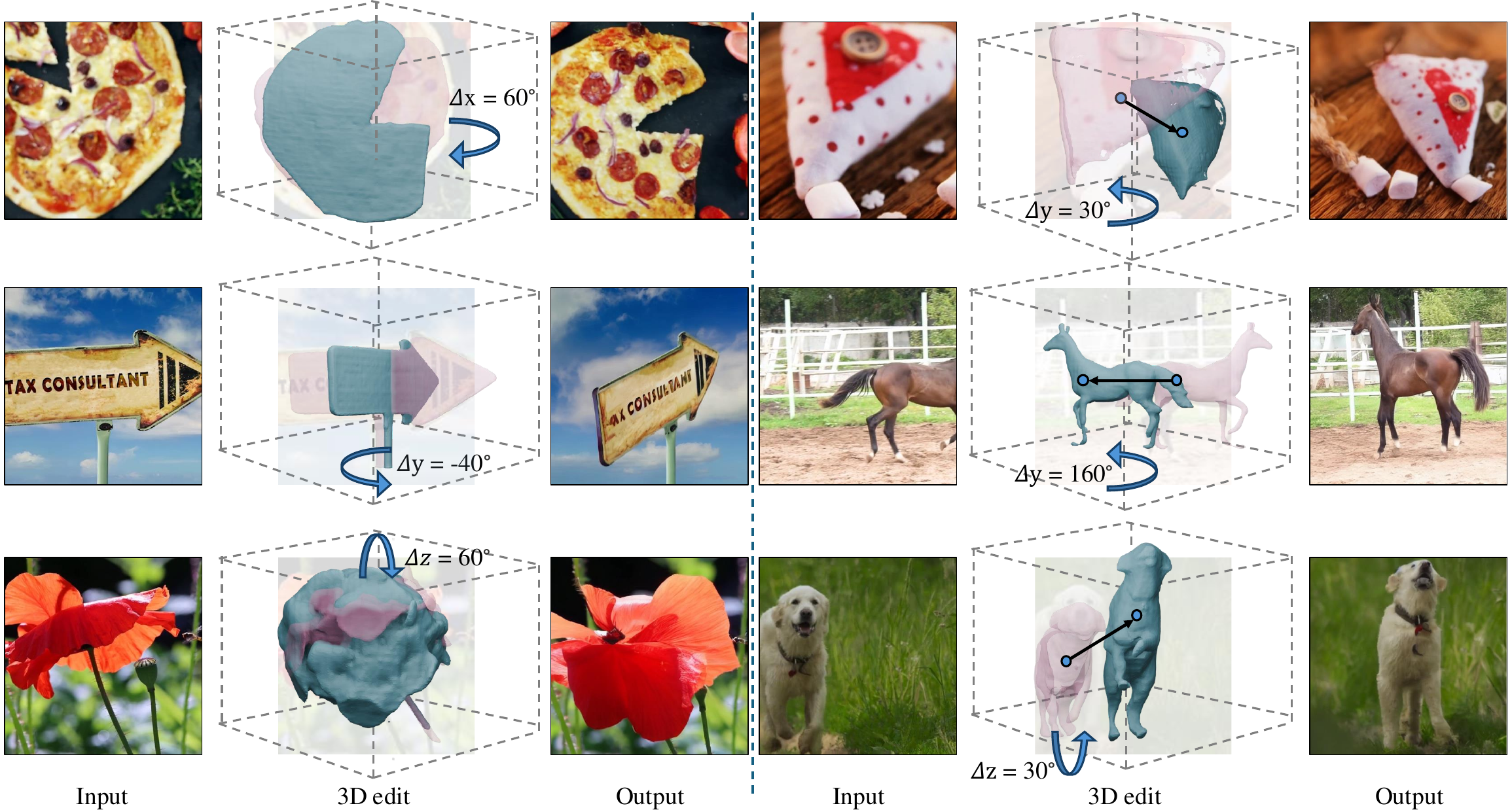}
  \caption{
        \textbf{{3D-aware photo editing. Given a source image with user-specified 3D transformations, our model generates a new image that follows the user's edit while preserving the input identity. \editI{The 3D edit is visualized via the transformation between the original mesh ({\color{pink}{pink}}) and the edited mesh ({\color{cyan}{cyan}}).}
        }
        }     
    }
  \label{fig:teaser}
\end{teaserfigure}

\maketitle

\input{1-intro}
\input{2-related_work}

\input{3-approach}

\input{4-experiments}
\input{5-conclusion}

\bibliographystyle{ACM-Reference-Format}
\bibliography{references}

\input{figs/3-fig-qual-ours}
\input{figs/3-fig-qual-comp-mf}

\end{document}

%% file: macro.tex
\definecolor{crimson}{rgb}{0.86, 0.08, 0.24}
\definecolor{gray}{rgb}{0.5,0.5,0.5}
\definecolor{green}{rgb}{0, 0.4, 0}
\definecolor{orange}{rgb}{1, 0.5, 0}
\definecolor{mahogany}{rgb}{0.75, 0.25, 0.0}
\definecolor{purple}{rgb}{0.6, 0, 0.6}
\definecolor{darkgreen}{rgb}{0, 0.4, 0}
\definecolor{frenchblue}{rgb}{0.0, 0.45, 0.73}
\definecolor{blue}{rgb}{0.0, 0.0, 0.65}
\definecolor{red}{rgb}{1,0,0}
\definecolor{yellow}{rgb}{1,1,0}
\definecolor{magenta}{rgb}{1,0,1}
\definecolor{pink}{rgb}{1,0.412,0.706}
\definecolor{cyan}{rgb}{0.25,0.55,0.70}
\definecolor{newgreen}{rgb}{0, 0.6, 0.2}

\newlength\paramargin
\newlength\figmargin
\newlength\subfigmargin
\newlength\subsecmargin
\newlength\tabmargin
\newlength\eqmargin

\newlength\presecmargin
\newlength\secmargin

\newlength\rulelength

%% file: 1-intro.tex
\section{Introduction}
\label{sec:intro}

Generative image editing is a growing research field that promises intuitive edits of objects in images, even if information about these objects or the scene that contains them is incomplete. For example, moving an object from one side of the image to the other, or rotating it as shown in Figure~\ref{fig:teaser}, may require knowledge about lighting, shadows, and occluded parts of the scene that are not available in the image. For these edits, generative models can hallucinate the missing information to obtain a plausible result. Current generative editing methods focus on appearance edits or 2D edits of image patches. However the objects we typically manipulate in images are projections of 3D objects. Some natural edits for 3D objects, like out-of-plane rotations or 3D translations, are thus not possible in most current approaches. However, 3D editing is crucial in applications such as e-commerce, where a 3D object may need to be shown from multiple angles, or digital media production, where it gives artists the ability to re-configure or relight a 3D scene shown in an image, without having to explicitly reconstruct the entire 3D scene, simplifying the creative process. By developing such a 3D editing method on natural images, we aim to bridge the gap between 2D and 3D workflows, making realistic edits more accessible for real-world applications.%

The current challenge for 3D editing of objects in images lies in maintaining consistent object appearance across different angles and lighting conditions, which is essential for creating seamless edits. Existing approaches have attempted to address these issues using either optimization or feed-forward deep net techniques. Optimization-based approaches, such as Image Sculpting~\cite{yenphraphai2024image}, begin by constructing a rough 3D shape of the object, followed by directly editing the 3D shape, and finally performing refinement to obtain the final edits. While this method achieves high-quality results, it is computationally intensive and slow, limiting its practical applications. %
In contrast, feed-forward approaches like 3DIT~\cite{michel2024object} and Magic Fixup~\cite{alzayer2024magic} leverage conditional diffusion models to guide the editing process, making them relatively fast and efficient. However, these methods are primarily limited by their dependence on 2D data and synthetic training sets, which either lack depth and spatial understanding or real-world data understanding. Besides, the reliance on text prompts restricts the precision and granularity of the user control, often leading to outputs that may diverge from the user's intention.

To address these challenges, we propose a feed-forward method that utilizes real-world video data enriched with 3D priors, allowing for realistic 3D-aware editing of objects in natural images.  We design a novel data generation pipeline to overcome the challenge of collecting large-scale 3D-aware image editing datasets in real-world scenarios. Our pipeline generates training data by leveraging 3D transformations estimated between frames in a video, and combines those with the priors obtained from an image-to-3D model. This intermediate 3D guidance serves as a crucial bridge, enabling the model to learn 3D-aware editing without requiring explicit 3D annotations for every frame. By utilizing both the dynamic information from videos and the structural insights provided by 3D priors, our approach captures real-world physical dynamics while facilitating fine-grained control over edits. This innovative design allows the model to generalize effectively to natural scenes, bridging the gap between synthetic and real-world applications. 
In Figure~\ref{fig:teaser}, we show some 3D-aware edits performed by our approach which allows fine-grained 3D user control while preserving object identity.%

Our contributions are threefold: (1) we develop a novel data pipeline for generating 3D-aware training data from real-world videos, bridging 2D inputs with 3D editing capabilities; (2) we design an efficient feed-forward model that performs precise 3D editing on natural images using this 3D guidance; and (3) we conduct extensive evaluations, demonstrating that our method achieves realistic 3D edits and outperforms state-of-the-art approaches.%

%% file: 2-related_work.tex
\section{Related Work}
\label{sec:related_work}

\subsection{3D-Aware Generative Image Editing}
3D-aware image editing methods provide 3D  control for image objects while maintaining consistency w.r.t.\ object identity, pose, and lighting during editing.
Object3DIT~\cite{michel2024object} is one of the earliest 3D-aware editing methods, directly operating on 3D transformation parameters as a condition in a feed-forward deep-net architecture. The method is however limited by its small synthetic training set, reducing its generality and the precision of its edits.
Diffusion Handles~\cite{pandey2024diffusionhandles}
and GeoDiffuser~\cite{sajnani2024geodiffuser}
are recent approaches that are general and precise, but use inference-time optimization to align the output to a particular edit, limiting  robustness and inference speed.
The space of supported 3D transformations is somewhat limited since those techniques
make no attempt in explicitly reasoning about unseen parts of objects.
In contrast, Image Sculpting~\cite{yenphraphai2024image} leverages off-the-shelf single-view reconstruction models to enable impressive 3D-aware editing results, but also requires a computationally demanding inference-time optimization. 
Unlike prior approaches, our work focuses on fine-tuning large image
diffusion models specifically for this task.
This allows our method to exhibit remarkable robustness to challenging editing operations
that could not be performed with existing baselines (see Figure~\ref{fig:3.qual.ours}). 
Additionally, no inference-time optimization is needed, allowing for fast evaluation. 

\subsection{Generative Image Editing}
Drag-based methods have emerged as a prominent paradigm for interactive image editing, offering users precise control over object movement and transformation. Early methods like DragGAN~\cite{pan2023drag} utilized GANs for point-to-point dragging but faced challenges with generalization and editing quality. More recent methods, such as DragDiffusion~\cite{mou2023dragondiffusion}, InstantDrag~\cite{shin2024instantdrag}, and EasyDrag~\cite{hou2024easydrag}, extend this concept to diffusion models, leveraging fine-tuned or reference-guided approaches to enhance photorealism. DragonDiffusion~\cite{mou2023dragondiffusion} stands out by avoiding fine-tuning and employing energy functions with visual cross-attention, enabling diverse editing tasks within and across images. DiffEditor~\cite{mou2024diffeditor} and DiffUHaul~\cite{avrahami2024diffuhaul} further refine drag-style editing, addressing challenges like entanglement and enhancing consistency in dragging results. For articulated object interactions, DragAPart~\cite{li2025dragapart} focuses on part-level motion understanding, allowing edits like opening drawers or repositioning parts. In contrast, generative editing methods that do not rely on drag-based interactions offer alternative workflows for tasks like object insertion, removal, and repositioning. ObjectDrop~\cite{winter2024objectdrop} models the effects of objects on scenes using counterfactual supervision, enabling realistic object manipulation. Meanwhile, SEELE~\cite{wang2024seele} formulates subject repositioning as a prompt-guided inpainting task, preserving image fidelity while offering precise spatial control.
Magic Fixup~\cite{alzayer2024magic} employs diffusion models to transform coarsely edited images into photorealistic outputs, leveraging video data to learn how objects adapt under various conditions. Similarly, ObjectStitch~\cite{song2023objectstitch} and IMPRINT~\cite{song2024imprint} focus on object compositing while preserving identity and harmonizing with the background, making them valuable for realistic image manipulation. %
However, unlike our approach, none of the methods in this paragraph benefit from a 3D-aware prior or provide controls to  support 3D-aware edits like out-of-plane rotations or 3D translations. %

\subsection{Image-to-video with motion control}
Image-to-video methods with motion control~\cite{wang2024boximator,guo2025sparsectrl,shi2024motion,bahmani2025tc4d} are related to generative image editing to some extent, as any generated frame could be taken as an edited image. However, edits are limited to motions that are plausible in a video and, to our best knowledge, none of the methods provide 3D control.

%% file: 3-approach.tex
\input{figs/2-fig_inference}
\input{figs/2-fig-overview-dp-part1}
\input{figs/2-fig-overview-dp-part2}
\input{figs/2-fig_data_exp}
\input{figs/2-fig_overview_model}

\section{Approach}
\label{sec:approach}
Our goal is 3D editing (e.g., out-of-plane rotation and translation) of a chosen object within an image. 
Existing 3D editing approaches that use inference-time optimization~\cite{pandey2024diffusionhandles, yenphraphai2024image} suffer from excessive inference times, making them impractical in real-world applications. In contrast, feedforward approaches~\cite{michel2024object} suffer from a lack of high-quality training data, limiting generality and control. %
We propose a feedforward 3D editing method that offers precise control and good generality. The editing workflow is illustrated in Figure~\ref{fig:2.inference}. As shown, the 3D edits we consider include out-of-plane rotations and translations that change the perspective of the object. 
Formally, given a source image $I_\mathrm{src}$, the user selects the object to be modified. The selection is represented via the mask $M_\mathrm{src}$, which we use to obtain a rough 3D reconstruction. %
The user then performs a  3D edit of the rough 3D reconstruction. Upon rendering the modified 3D reconstruction we obtain the 
3D guidance $I_\mathrm{guide} \in \mathbb{R}^{H \times W \times 3}$, which is used to generate the desired editing result.

To provide the necessary supervision for training the feedforward model, i.e., to obtain a source image $I_\mathrm{src}$, a guidance image $I_\mathrm{guide}$, and a ground truth target image $I_\mathrm{tgt}$, we construct a new dataset derived from videos. For this, we develop the data processing pipeline %
that we describe in Section~\ref{sec:dataset_construction}. %
As videos naturally capture 3D motion as well as variations in lighting and background conditions, they offer a rich source of real-world data. By integrating this dataset into the training process, our method learns to handle complex 3D transformations while ensuring photorealism and maintaining the fidelity of the edited subject.

Using this dataset, we fine-tune a pretrained diffusion model conditioned on 1) the  edited guidance image $I_\mathrm{guide}$, and 2) the source image $I_\mathrm{src}$. Importantly, unlike prior 2D editing methods, the guidance image is obtained by 3D-transforming a full 3D reconstruction of the chosen object in the image. We describe architecture and training setup of this model in Section~\ref{sec:diffusion_model}. %

\input{3.1-data_pipeline}

\input{3.2-model}

\input{3.3-implement_detail}

%% file: figs/2-fig_inference.tex
\begin{figure}[t]
    \centering
    \includegraphics[width=\linewidth]{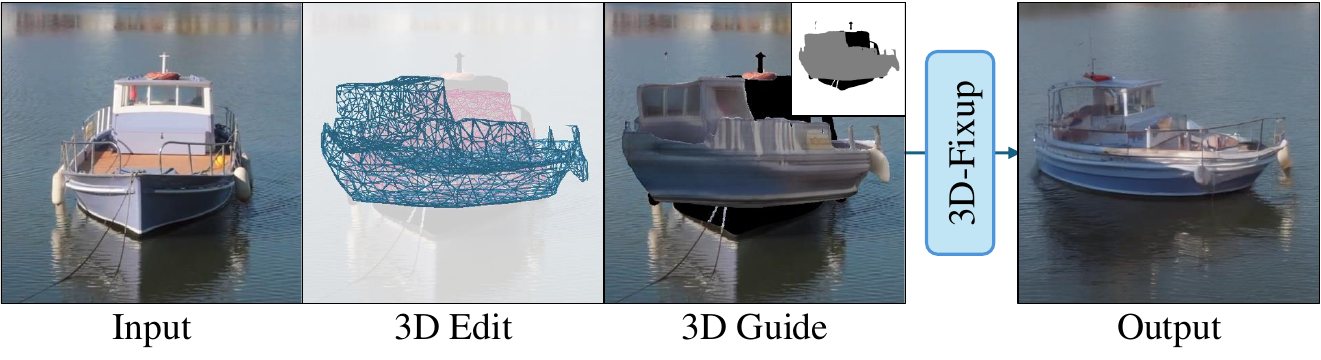}
    \caption{
    \textit{Inference pipeline.} %
    We assume editing instructions (possibly converted from text prompts) are in the form of 3D operations like rotation and translation. Given a mask indicating the object to be edited, we first perform image-to-3D~\cite{xu2024instantmesh} to reconstruct the mesh. We then apply the user's desired 3D edit to obtain the 3D guidance. \editI{Here the 3D edit is visualized as the transformation between original mesh ({\color{pink}{pink wire-frame}}) and the edited mesh ({\color{cyan}{cyan wire-frame}})}. Finally, the model outputs the 3D aware editing result.
    }
    \vspace \figmargin
    \label{fig:2.inference}
\end{figure} 

%% file: figs/2-fig-overview-dp-part1.tex
\begin{figure*}[t]
    \centering
    \includegraphics[width=\linewidth]{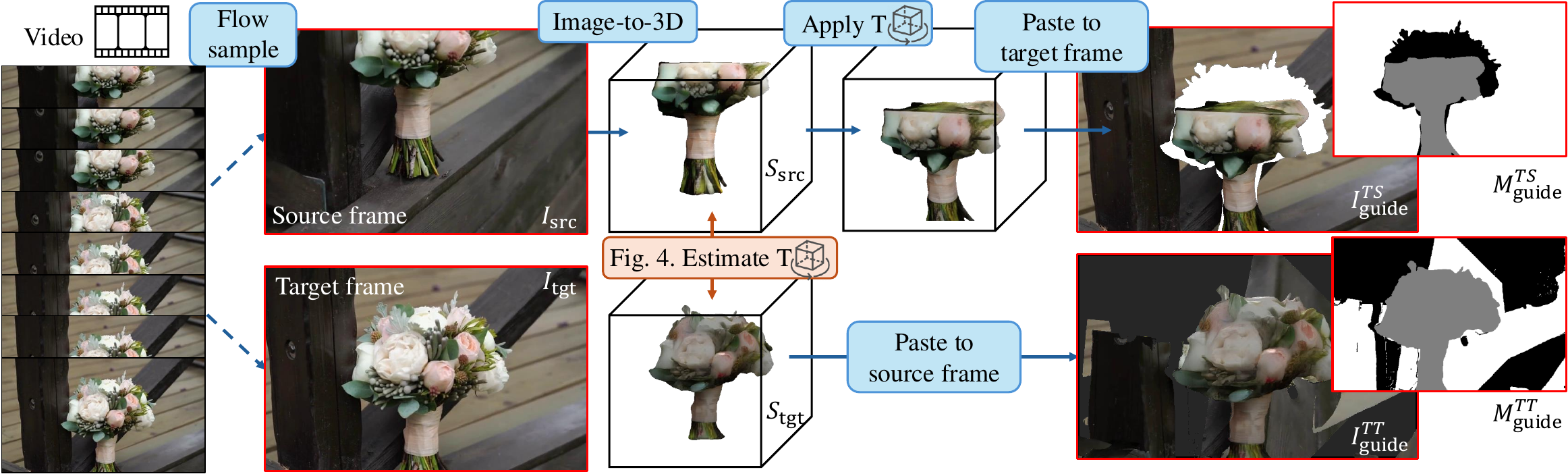}
    \caption{
     \textit{Data pipeline: Overview.} Given a video, we sample two frames, the source frame $I_{\mathrm{src}}$ and the target frame $I_{\mathrm{tgt}}$, using optical flow as a cue: we discard videos where the flow indicates little motion through the entire clip. Using Image-to-3D methods, we reconstruct a mesh for the desired object for both frames. We then estimate the 3D transformation $\mathbf{T}$ (see Figure~\ref{fig:2.overview.dp2}) between the source frame mesh and the target frame mesh. Availability of the transformation $\mathbf{T}$ %
     enables two ways 
     to create the training data: (1) in ``Transform Source'', we paste the rendering of the transformed source mesh onto the target frame; (2) in ``Transform Target'', we paste the rendering of the target mesh onto the source frame. Data examples are shown in Figure~\ref{fig:2.dataexp}.
    }
    \vspace \figmargin
    \label{fig:2.overview.dp1}
\end{figure*} 

%% file: figs/2-fig-overview-dp-part2.tex
\begin{figure}[t]
    \centering
    \includegraphics[width=\linewidth]{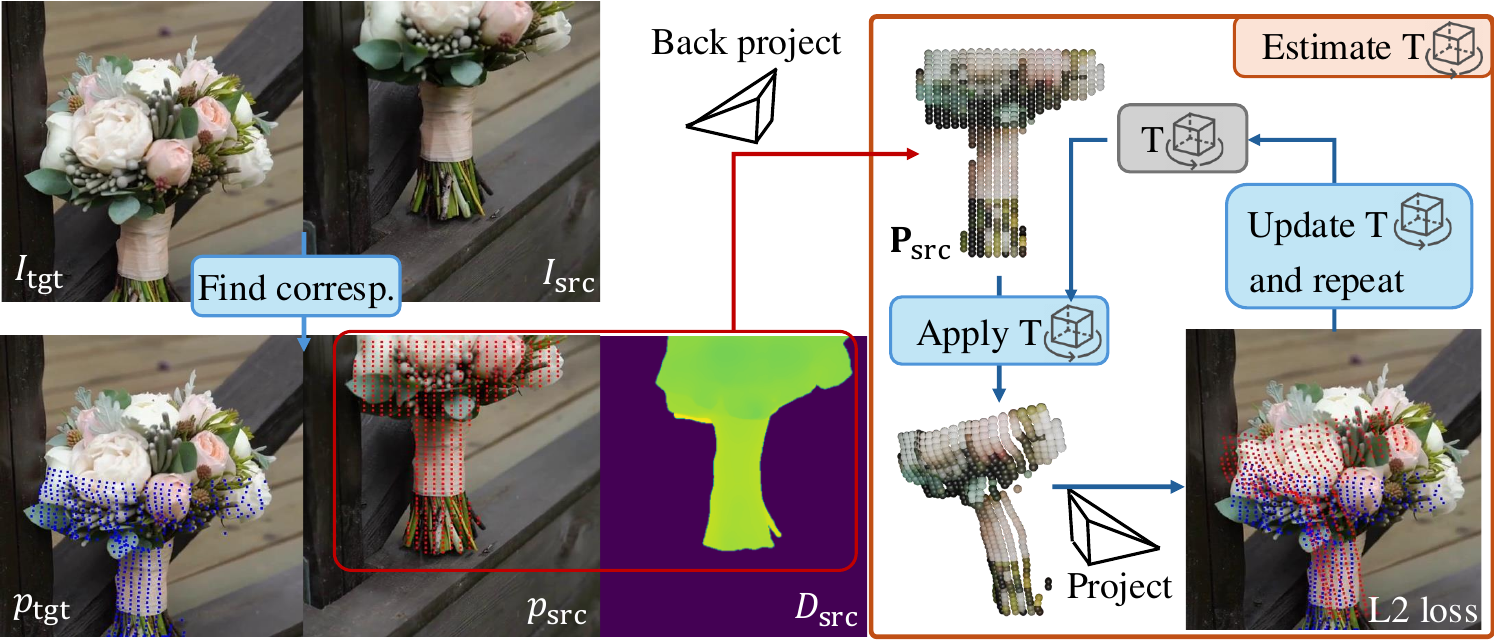}
    \vspace{-5mm}
    \caption{
    \textit{Data pipeline: Estimation of the 3D transformation $\mathbf{T}$.} We estimate the 3D transformation $\mathbf{T}$ by leveraging  correspondences between the source frame and the target frame. Given  frames between  the source frame and the target frame, we first perform tracking to obtain corresponding points. We then initialize the parameters for the 3D transformation $\mathbf{T}$ and use an optimization  to  improve $\mathbf{T}$: (1) We unproject the 2D correspondences on the source frame to 3D pointclouds and %
    apply the current $\mathbf{T}$ to transform points to the target image; (2) we project points back to 2D and compare via an L2 loss with the 2D correspondences of the target frame. %
    }
    \vspace{-5mm}
    \vspace \figmargin
    \label{fig:2.overview.dp2}
\end{figure} 

%% file: figs/2-fig_data_exp.tex
\begin{figure}[t]
    \centering
    \vspace{\figmargin}
    \includegraphics[width=\linewidth]{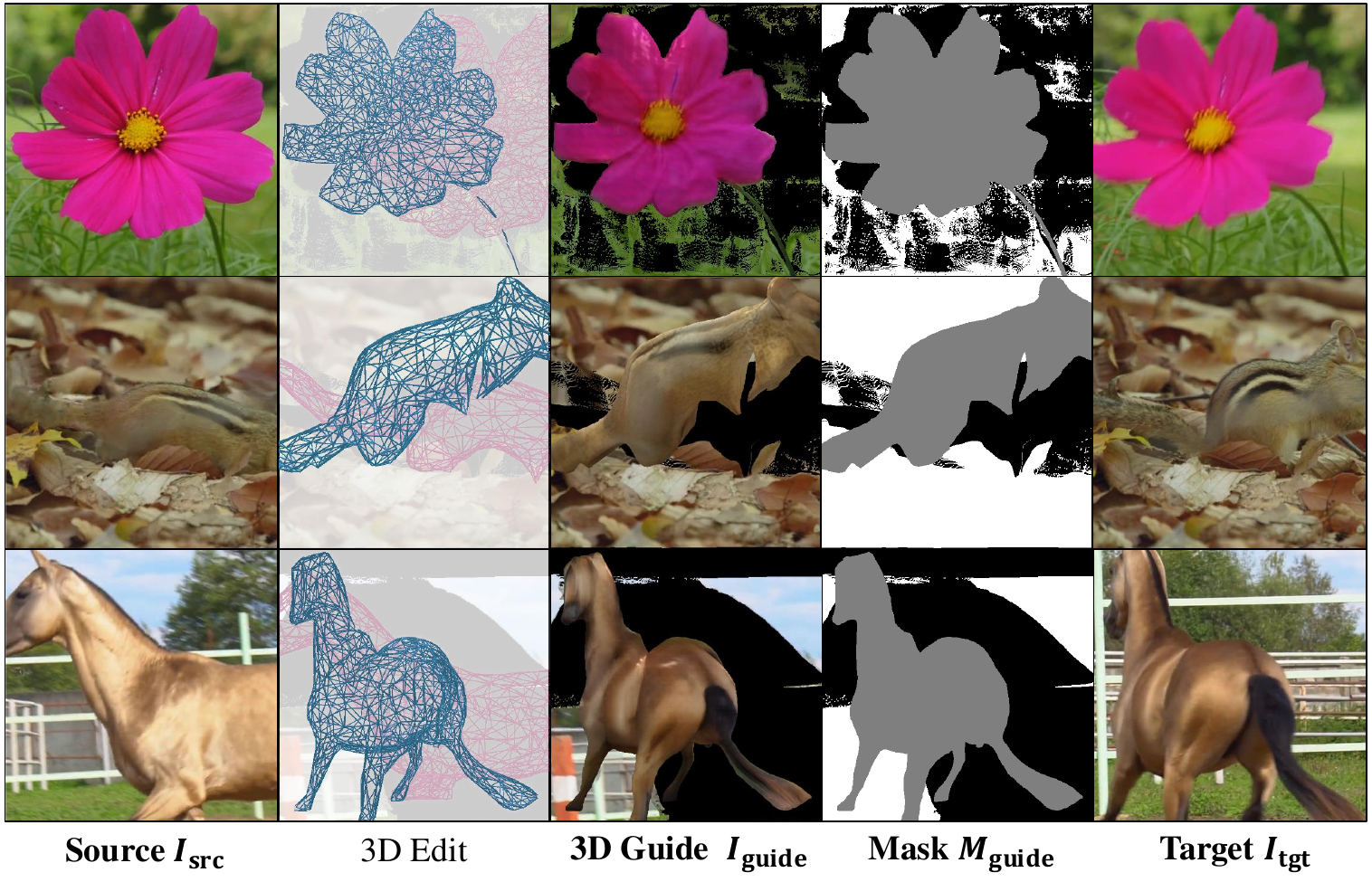}
    \caption{
    \textit{Examples of the training data.} 
    Given a video, we use the steps described in Figure~\ref{fig:2.overview.dp1} to obtain the training data, i.e., the source image $I_{\mathrm{src}}$ and the target image $I_{\mathrm{tgt}}$. The guidance image is obtained via the developed data pipeline. The mask has three values: 0 indicates the hole created by the coarse edit and the model needs to inpaint by looking at the details of the reference; 0.5 refers to the rendering of the object; and 1.0 denotes the original background.
    }
    \vspace \figmargin
    \label{fig:2.dataexp}
\end{figure} 

%% file: figs/2-fig_overview_model.tex
\begin{figure}[t]
    \centering
    \includegraphics[width=\linewidth]{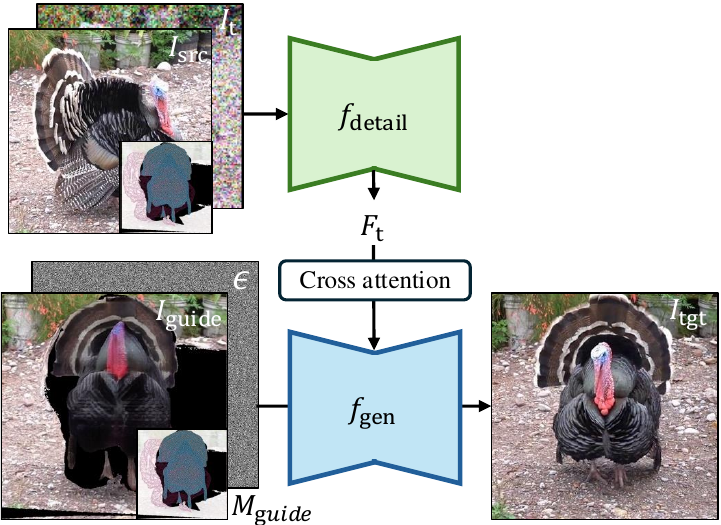}
    \caption{
    \textit{Overview of the training pipeline}. %
    We develop a conditional diffusion model for 3D-aware image editing. \editI{It consists of two networks: $f_{\mathrm{gen}}$ and $f_{\mathrm{detail}}$. During training, given the inputs—target frame $I_{\mathrm{tgt}}$, 3D guidance $I_{\mathrm{guide}}$, mask $M_{\mathrm{guide}}$, and detail feature $F_t$—$f_{\mathrm{gen}}$ learns the reverse diffusion process to predict the noise $\epsilon$ and reconstruct $I_{\mathrm{tgt}}$. To better preserve identity and fine-grained details from the source image $I_{\mathrm{src}}$, $f_{\mathrm{detail}}$ takes as input the source image $I_{\mathrm{src}}$, its noisy counterpart $I_t$, and the mask $M_{\mathrm{guide}}$, and extracts detail features $F_t$. We apply cross-attention between $F_t$ and the intermediate features of $f_{\mathrm{gen}}$ to incorporate content and details from $I_{\mathrm{src}}$ during the reverse diffusion process.}
    }
    \vspace \figmargin
    \label{fig:2.overview.model}
\end{figure}

%% file: 3.1-data_pipeline.tex
\subsection{Constructing the Dataset from Videos}
\label{sec:dataset_construction}
Given a video, we create data pairs by sampling two frames, a source image $I_{\mathrm{src}}$ and a target image $I_{\mathrm{tgt}}$. We use both images to compute a guidance image $I_\mathrm{guide}$. 
Figure~\ref{fig:2.overview.dp1} provides an overview of our data processing pipeline. %
We discuss details next.

\paragraph{\textbf{Flow for sampling the source and target image.}} We compute optical flow across all frames in a given video. If the accumulated flow across the entire video is too small,  we discard the video. If the accumulated flow exceeds a threshold, we sample two frames from the video clip which we refer to as $I_{\mathrm{src}}$ and $I_{\mathrm{tgt}}$.

\editI{\paragraph{\textbf{Obtaining mask for the main object.}} We use Grounded-SAM~\cite{ren2024grounded} to obtain object masks and filter cases with privacy, aesthetic, or insignificant 3D transformation issues. Notice that Grounded-SAM struggles to detect occluded or unusually shaped instances, which are rare. To automatically identify the main object, we define an ``instance score'' that prioritizes centrally located objects that occupy significant portions of the frame. The score is the weighted sum of inverted border score $S_{b}$ and area score $S_{a}$ across the video. They are calculated by
\editII{
\begin{align}
S_{b} = 1 - \frac{p_{\text{border}}^{\text{inst}}}{p_{\text{border}}}, \quad S_{a}=\frac{p_{\text{inst}}}{p_{\text{image}}},
\end{align}
}
where $p_{\text{border}}$ is the number of border pixels in the whole image, $p_{\text{border}}^{\text{inst}}$ is the number of instance pixels which touch the border, $p_{\text{inst}}$ is the total number of pixels of the instance, and $p_{\text{image}}$ is the number of pixels of the whole image.  The instance score is $S_{\text{inst}} = 0.6 * S_{b} + 0.4 * S_{a}$. We select the highest-scoring instance. 
}
\paragraph{\textbf{Image to 3D reconstruction.}} Given the source image $I_{\mathrm{src}}$ and the target image $I_{\mathrm{tgt}}$, we perform image to 3D reconstruction using InstantMesh~\cite{xu2024instantmesh}.
Specifically, given the foreground masks $M_{\mathrm{src}} \in \mathbb{R}^{H \times W}$ and $M_{\mathrm{tgt}} \in \mathbb{R}^{H \times W} $, we first crop the foreground object and pad the image to a square while ensuring that the object is centered. InstantMesh then employs \editI{Zero123++~\cite{shi2023zero123plus}} to generate multiview images. Subsequently, InstantMesh operates on the multiview images to compute the reconstructed meshes $S_{\mathrm{src}}$ and $S_{\mathrm{tgt}}$ for the source and target images. 

\noindent\paragraph{\textbf{Estimating 3D transformation with tracking.}} To obtain a coarse edit in an automated manner, we use the  object meshes for the source and target image to estimate a 3D transformation. This process is illustrated in Figure~\ref{fig:2.overview.dp2}. 
Concretely, to estimate the 3D transformation $\mathbf{T}$ between the object in the source image $I_{\mathrm{src}}$ and the target image $I_{\mathrm{tgt}}$, we first compute $N$ correspondences $p_{\mathrm{src}}$ and $p_{\mathrm{tgt}}$ along with their visibility maps $v_{\mathrm{src}}$ and $v_{\mathrm{tgt}}$, using %
SpaTracker~\cite{xiao2024spatialtracker}. The input to the tracking model is the video segment containing the two frames $I_{\mathrm{src}}$ and $I_{\mathrm{tgt}}$, along with the source mask $M_{\mathrm{src}}$. Depth maps $D_{\mathrm{src}}$ and $D_{\mathrm{tgt}}$ are rendered from the meshes $S_{\mathrm{src}}$ and $S_{\mathrm{tgt}}$, respectively. 

Using the depth maps and correspondences, we backproject the 2D points into 3D space:
\begin{align}
    \mathbf{P}_{\mathrm{src}} &= \Pi^{-1}(p_{\mathrm{src}}, D_{\mathrm{src}}, \mathbf{K}), \\
    \mathbf{P}_{\mathrm{tgt}} &= \Pi^{-1}(p_{\mathrm{tgt}}, D_{\mathrm{tgt}}, \mathbf{K}),
\end{align}
where $\Pi^{-1}$ denotes the unprojection operation and $\mathbf{K} \in \mathbb{R}^{3 \times 3}$ is the intrinsic camera matrix. The 3D points $\mathbf{P} \in \mathbb{R}^3$ are calculated as:
\begin{align}
    \mathbf{P} = D(p) \cdot \mathbf{K}^{-1} \begin{bmatrix} p_x \\ p_y \\ 1 \end{bmatrix},
\end{align}
where $D(p)$ is the depth at pixel $p = (p_x, p_y)$.

The translation component of $\mathbf{T}$ is initialized by calculating the centroid offset between the two point clouds:
\vspace{\eqmargin}
\begin{align}
    \mathbf{t} = \mathbf{c}_{\mathrm{tgt}} - \mathbf{c}_{\mathrm{src}}, \quad \mathbf{c} = \frac{1}{N} \sum_{i=1}^N \mathbf{P}_i.
\end{align}
\vspace{\eqmargin}

To initialize the rotation, a coarse grid search is performed jointly over the $X$, $Y$, and $Z$ axes, using a step size of $10^\circ$ within the range $[0^\circ, 360^\circ]$. The transformation $\mathbf{T}$ is optimized by minimizing the re-projection loss:
\begin{align}
    \mathcal{L}_{\mathrm{reproj}} = \sum_{i=1}^N \| p_{\mathrm{tgt},i} - \Pi(\mathbf{T} \mathbf{P}_{\mathrm{src},i}, \mathbf{K}) \|_2^2,
    \label{eq:repoj.loss}
\end{align}
where $\Pi$ is the projection operation:
\begin{align}
    \Pi(\mathbf{P}, \mathbf{K}) = \begin{bmatrix} \mathbf{K} & \mathbf{0} \end{bmatrix} \mathbf{P}.
\end{align}
The optimization iteratively adjusts $\mathbf{T}$ until convergence. 
\editI{Eq.~\ref{eq:repoj.loss} assumes $\mathbf{T}$ to be rigid transformations. However, for non-rigid transformations, Eq.~\ref{eq:repoj.loss} finds a close rigid approximation. During training, the model learns from non-rigid transformation as ground truth while using a rigid approximation in guidance $I_\text{guide}$. This discrepancy is often desirable, as it encourages the model to adapt non-rigidly, ensuring the edited object fits naturally into its new context. For example, when rotating the dog in Fig.~\ref{fig:teaser} or the horse in Fig.~\ref{fig:3.qual.comp.mf}, subtle posture adjustments, such as foot placement, help the resulting scene remain plausible.} %
\paragraph{\textbf{Creating the training data.}} With the optimized transformation $\mathbf{T}$ computed, we have two settings to obtain the guidance image and the editing mask. We refer to the first setting as ``Transform Source'' (TS): the estimated 3D transformation $\mathbf{T}$ is applied to the source mesh $S_{\mathrm{src}}$. The transformed mesh is rendered and pasted onto the target image $I_{\mathrm{tgt}}$ based on the target mask $M_{\mathrm{tgt}}$ to form the guidance image $I_{\mathrm{guide}}^{TS}$. \editI{The editing mask $M_{\mathrm{guide}}^{TS}$ has $1.0$ for the static background, $0.5$ for the rendered regions, and $0.0$ for the holes created by cropped the object in the $I_{\mathrm{tgt}}$}. We refer to the second setting as ``Transform Target'' (TT): we transform and render the target mesh $S_{\mathrm{tgt}}$ onto the source frame $I_{\mathrm{src}}$ to obtain the guidance image $I_{\mathrm{guide}}^{TT}$. \editI{The background is warped based on the flow computed between the source frame and the target frame following Alzayer~\etal~\cite{alzayer2024magic}. $M_{\mathrm{guide}}^{TT}$ is similar to $M_{\mathrm{guide}}^{TS}$ for the background ($1.0$) and rendered regions ($0.5$), while the holes $(0.0)$ indicated the warped regions.} Thus, the final training tuple for ``Transform Source'' and ``Transform Target'' are $(I_{\mathrm{source}}, I_{\mathrm{guide}}^{TS}, M_{\mathrm{guide}}^{TS}, I_{\mathrm{target}})$, and $(I_{\mathrm{source}}, I_{\mathrm{guide}}^{TT}, M_{\mathrm{guide}}^{TT}, I_{\mathrm{target}})$ respectively. \editI{We sample data pair randomly from TT or TS when training the model. Both the editing masks $M_{\mathrm{guide}}^{TS}$ and $M_{\mathrm{guide}}^{TT}$ guide the model to inpaint missing regions (0.0), enhance 3D-transformed areas (0.5), and preserve intact content (1.0).} Examples of the collected data are illustrated in Figure~\ref{fig:2.dataexp}.

\editI{Our dataset is sourced from 8 million licensed videos. We discard videos exceeding 500 frames due to high compute costs, reducing the dataset to 2 million. After filtering cases depicting humans (privacy and aesthetics issues),  illustrations and drone videos (insignificant 3D transformations), and videos without detected entities, we retain 375k clips. Further motion-based flow filtering reduces this to 50k videos. Note that the proposed data pipeline is general and applicable to any video dataset. 
Processing each video takes $\sim$95.7s and requires 14.41GB of memory on a single A100 GPU. The pipeline includes flow filtering, mask extraction, image-to-3D, tracking, and 3D estimation, with videos averaging 50–500 frames.}

%% file: 3.2-model.tex
\subsection{3D Editing with a Diffusion Model}
\label{sec:diffusion_model}

Our diffusion model aims to generate realistic images that complete the 3D edit specified by the guidance image $I_{\mathrm{guide}}$. Hence, the structure of the image should be preserved as outlined in the guidance. For regions indicated by the mask $M_{\mathrm{guide}}$, the model performs the following operations: inpainting missing regions ($M_{\mathrm{guide}} = 0.0$), modifying ambiguous regions ($M_{\mathrm{guide}} = 0.5$), and preserving content and identity in confident regions ($M_{\mathrm{guide}} = 1.0$).

An overview of our architecture is provided in Figure~\ref{fig:2.overview.model}. We base our architecture on MagicFixup~\cite{alzayer2024magic} and adopt two networks in our pipeline: a generator for generating the output image $f_{\mathrm{gen}}$ and an extractor for extracting details $f_{\mathrm{detail}}$ from the source image $I_{\mathrm{src}}$. We use those networks in a diffusion process which operates as follows: the diffusion forward process progressively adds Gaussian noise to an image, yielding a sequence of intermediate states:
\begin{align}
x_t \sim \mathcal{N}(\sqrt{\alpha_t} x_{t-1}, (1 - \alpha_t) \mathbf{I}),
\end{align}
which gradually resemble a standard Gaussian as diffusion time $t$ increases towards a final timestep $T$. Here $\alpha_t$ defines the noise schedule at diffusion time $t$. The model learns the reverse process: a standard Gaussian input $x_T \sim \mathcal{N}(0, \mathbf{I})$ is gradually denoised toward intermediate states $x_t$ before eventually arriving at the final estimated image $x_0$:
\editI{
\begin{align}
x_{t-1} = f_{\mathrm{gen}}(x_t, I_{\mathrm{guide}}, M_{\mathrm{guide}}, F_t, t; \theta).
\end{align}
At each denoising step $t$, the model is conditioned on the guidance $I_{\mathrm{guide}}$, mask $M_{\mathrm{guide}}$, and the features $F_t$ extracted by the extractor $f_{\mathrm{detail}}$.
We initialize the process from a noisy version of the  guidance image, i.e., we use
\begin{align}
x_T = \sqrt{\bar{\alpha}_T} I_{\mathrm{guide}} + \sqrt{1 - \bar{\alpha}_T} \epsilon,
\end{align}
where $\epsilon \sim \mathcal{N}(0, \mathbf{I})$ and $\bar{\alpha}_T = \prod_{s=1}^T \alpha_s$. This initialization ensures alignment with the guidance while bridging the gap between training and inference domains.
The extractor $f_{\mathrm{detail}}$ operates on the source image $I_{\mathrm{src}}$ for referencing and with the goal to preserve fine-grained details and object identity. To ensure compatibility with the diffusion process, we add noise to the source image:
\vspace{\eqmargin}
\begin{align}
I_t = \sqrt{\bar{\alpha}_t} I_{\mathrm{src}} + \sqrt{1 - \bar{\alpha}_t} \epsilon, \quad\!\! \epsilon \sim \mathcal{N}(0, \mathbf{I}), \quad\!\! \bar{\alpha}_t = \prod_{s=1}^t \alpha_s.
\end{align}
\vspace{\eqmargin}
From this noisy source image, the extractor computes the feature 
\vspace{\eqmargin}
\begin{align}
F_t = [f_t^1, \ldots, f_t^n] = f_{\mathrm{detail}}([I_t, I_{\mathrm{src}}, M_{\mathrm{guide}}]; t),
\end{align}
\vspace{\eqmargin}
for each self-attention block. Here $n$ is the number of attention blocks, and $[\cdot]$ denotes concatenation along the channel dimension. These features are injected into the model through cross-attention layers, enabling details preservation from the source image to outputs during synthesis. For each layer $i$, at step $t$, the features $F_t$ extracted by the extractor $f_{\mathrm{detail}}$ serve as keys $K$ and values $V$, while the features of the generator $f_{\mathrm{gen}}$ $[g_t^1, \ldots, g_t^n]$ act as queries $Q$. Formally, we have
\vspace{\eqmargin}
\begin{align}
A_t^i = \mathrm{softmax}( \frac{Q_t^i K_t^{i\top}}{\sqrt{d}} ), \quad\text{and}\quad 
G_t^i = A_t^i V_t^i,
\end{align}
\vspace{\eqmargin}
where $Q_t^i$, $K_t^i$, and $V_t^i$ are query, key, and value projections of the respective features. This mechanism ensures that fine details from $I_{\mathrm{src}}$ are faithfully transferred to the synthesized output.
}

During inference, user instructions (e.g., text prompts, drags on 3D objects) are converted into 3D transformations $\mathbf{T}$ (out-of-plane rotations and translations). Using InstantMesh~\cite{xu2024instantmesh}, we perform image-to-3D reconstruction to generate a 3D mesh of the subject. Applying $\mathbf{T}$ to the mesh, we obtain the guidance for editing. The model uses this guidance $I_{\mathrm{guide}}$ along with the mask $M_{\mathrm{guide}}$ to produce the final output. Figure~\ref{fig:2.inference} illustrates this framework.

%% file: 3.3-implement_detail.tex
\subsection{Implementation details}
\editI{For fair comparisons, following the state-of-the-art Magic-Fixup~\cite{alzayer2024magic}, we train our model starting from pretrained weights of Stable Diffusion 1.4. Training samples are drawn from data settings—TT, TS, MF—with probabilities (0.35, 0.35, 0.3), where MF is sampling from Magic Fixup's data. To encourage identity preservation, we drop the conditioning on $I_\text{src}$ with a $0.2$ probability, forcing the model to rely on $I_\text{src}$'s context.}
We train the model with a batch size of 8, using AdamW~\cite{loshchilov2017decoupled} and a learning rate of $1\mathrm{e}{-5}$ on 8 NVIDIA A100 GPUs for about two days. We use a linear diffusion noise schedule, with $\alpha_1 = 0.9999$, $\alpha_T = 0.98$, and $T = 1000$. We use DDIM for sampling with 100 steps during inference time. The images were all cropped to $512 \times 512$ for training.

%% file: 4-experiments.tex
\input{figs/3-fig-qual-comp-baselines}

\section{Experiments}
\label{sec:exp}

We evaluate the proposed method both qualitatively and quantitatively on a set of edits. For this, we curated a set of user edits to show the use cases of the model in practical applications. We also created a test dataset which contains large 3D transformations to validate the proposed method. 

\paragraph{\textbf{Dataset.}} We use stock video as our dataset for training and testing. The training data consists of around 50k data points which contains common objects with motions in the scene. \editI{We randomly sample diverse scenes and objects while maintaining a reasonable scale when constructing the test set.}

\paragraph{\textbf{Baseline.}}
\editII{To validate the effectiveness of the proposed method, we compare to seven baselines: \textit{Magic Fixup}~\cite{alzayer2024magic}, \textit{Object 3DIT}~\cite{michel2024object}, \textit{Zero-1-to-3}~\cite{liu2023zero}, \textit{InstantDrag}~\cite{shin2024instantdrag}, \textit{MOFA-Video}~\cite{niu2024mofa}, \textit{Blended LDM}~\cite{avrahami2023blended}, and finally \textit{Instruct-pix2pix}~\cite{brooks2023instructpix2pix}. Please refer to the supplementary materials for details regarding the baselines.
}

\paragraph{\textbf{Metrics.}} For quantitative evaluation, we use LPIPS~\cite{zhang2018unreasonable} and FID~\cite{heusel2017gans} metrics. LPIPS measures the perceptual similarity to assess fidelity to the ground truth using a neural network such as AlexNet~\cite{krizhevsky2012imagenet} or VGGNet~\cite{simonyan2014very}. FID (Fréchet Inception Distance) evaluates the realism of generated images by comparing their distribution to that of real data.

\vspace{-2mm}
\subsection{Comparison with Baselines}
\editI{We compare the proposed method with several state-of-the-art image-editing methods, which operate on different types of conditions, such as the 3D transform, a drag (point correspondence), and a text prompt. The results are shown in Figure~\ref{fig:3.qual.comp.base}.
Similar to our approach, 3DIT~\cite{michel2024object} and Zero123~\cite{liu2023zero} use the 3D transform as condition. However, 3DIT fails to generate plausible results because of the domain gap between its synthetic training data and real-world images, while Zero123 struggles with identity preservation.
For the dragging-based methods InstantDrag~\cite{shin2024instantdrag} and MOFA-Video~\cite{niu2024mofa}, we use the known correspondence between the source and transformed mesh to define an input drag. We find that drags are too ambiguous to clearly define a 3D transform and both methods struggle to interpret larger drags, such as the rotation of the goldfish or the shoes.
Blended LDM~\cite{avrahami2023blended} takes the guidance image and the mask as inputs and adopts Blended Diffusion~\cite{avrahami2022blended} to refine the coarse edit, which does not preserve identity.
Finally, Instruct-pix2pix~\cite{brooks2023instructpix2pix} is instructed by a text prompt, but suffers from its ambiguity.
In contrast, our proposed method can generate high-quality edits for large 3D transformations while preserving identity.}

We also present a comparison with Magic Fixup in Figure~\ref{fig:3.qual.comp.mf}. The results demonstrate that our proposed method achieves more realistic images, benefiting from 3D-transformation-based guidance. For example, our method effectively handles pose changes, such as adjusting the camera's viewing direction or modifying the poses of subjects, \editII{as shown in the horse, jaguar, and cake examples.}
In contrast, Magic Fixup struggles with such edits. 

\subsection{3D Editing with Continuous Rotations} We also demonstrate that the proposed method can handle extensive 3D edits on common objects, as illustrated in Figure~\ref{fig:3.qual.ours}. In each scenario, we progressively increase the rotation from 0 to 180 degrees along the y-axis, applying it to the reconstructed mesh to generate the 3D-transformation-based guidance image. The results show that our method successfully deals with out-of-plane 3D rotation edits, from minor adjustments to substantial transformations, highlighting the model's 3D-awareness during editing.

\input{tables/tab_quan_comp}

\subsection{Quantitative Comparison of 3D Editing} We compute metrics to evaluate the performance of methods as shown in Tab.~\ref{tab:quan.comp}. LPIPS is calculated for each model by measuring the similarity between its outputs and the ground-truth images. The final LPIPS score is obtained as the mean value across all pairs. FID assesses the realism of the generated results by comparing the distribution of the generated images to that of real video frames. The results demonstrate that the proposed method produces outputs that are highly realistic and align well with the real data as indicated by lower FID and LPIPS values.
\editI{In terms of the detail preservation, we address the challenging task of hallucinating novel views and missing parts based on $I_\text{guide}$. Since it relies only on single-view image to generate unseen regions, preserving identity and fine details is inherently difficult. However, LPIPS in Table~\ref{tab:quan.comp} shows that 3D-Fixup achieves better detail preservation than prior methods.}
Finally, we also compare the runtime with Image-sculpting~\cite{yenphraphai2024image}, an optimization-based method for image editing. The runtime is $\sim877s$ per sample, while ours \editI{can achieve $\sim2s$ for 50 DDIM steps}.

\input{tables/tab_quan_ablation_data_comp}
\input{tables/tab_quan_ablation_cond}

\subsection{Ablation Study of Data Setting and Conditioning} We evaluate the importance of each data setting in Tab.~\ref{tab:quan.ablation}. Using the setting with all the data yields the best performance in terms of FID and LPIPS. \editI{We also evaluate the effect of different conditioning mechanisms in Tab.~\ref{tab:quan.ablation.cond}. First, we consider different mask settings: $(0.0,0.5,1.0)$ vs. $(0.0,1.0)$. Then we study the impact of dropout of image features for cross-attention. The results suggest that our conditioning outperforms other configurations.}

%% file: figs/3-fig-qual-comp-baselines.tex
\begin{figure*}[htbp]
    \centering
    \includegraphics[width=\linewidth]{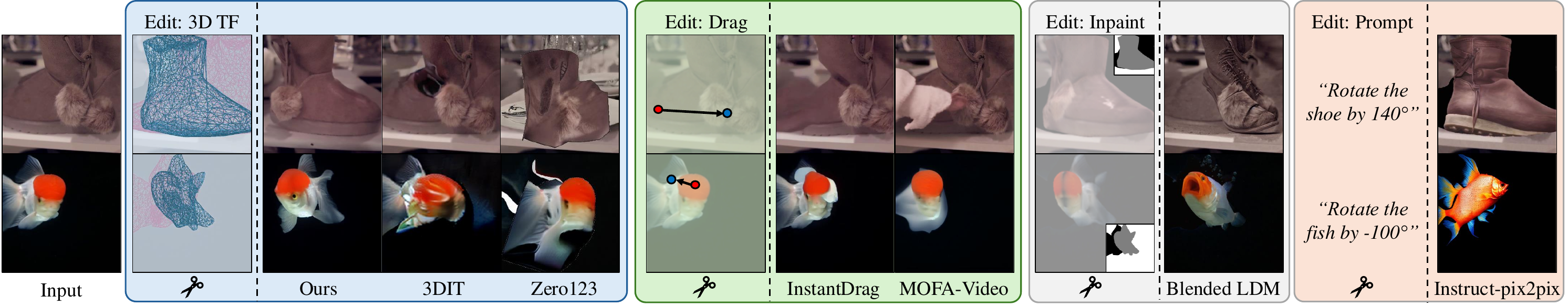}
    \vspace{-4mm}
    \caption{
    \textbf{Comparison with baselines.} \editI{We compare several state of the art baselines with different kinds of conditions, such as 3D transforms, drags, inpainting masks, and text prompts. We can see that none of the baselines accurately follow the target 3D transform while preserving identity. Baselines that directly use 3D transforms suffer from a lack of good training data, and using other types of conditions makes it hard to unambiguously specify the 3D transform.}
    }
    \vspace \figmargin
    \vspace{-3mm}
    \label{fig:3.qual.comp.base}
\end{figure*}

%% file: tables/tab_quan_comp.tex
\begin{table}[t]
\caption{
    \textbf{Quantitative comparison to baselines}. \editI{We compare to the baselines using the  LPIPS and FID metrics. The result shows that the 3D editing of the proposed method is closer to the ground truth and real distribution.}
}
\vspace{\tabmargin}
\centering
\setlength{\tabcolsep}{0.25em}
\small
\begin{tabular}{rr cc cc cc } 
    \toprule
    Model & LPIPS $\downarrow$ & FID (5k) $\downarrow$ & FID (30k) $\downarrow$ \\
    \midrule
    Magic Fixup~\cite{alzayer2024magic} & 0.5776 & 174.6926 & 27.1542 \\
    3DIT~\cite{michel2024object} & 0.4493 & 145.3389 & 23.8392 \\
    Zero123~\cite{liu2023zero} & 0.6803 & 202.2304 & 44.8570 \\
    Instruct-pix2pix~\cite{brooks2023instructpix2pix} & 0.7532 & 231.7562 & 73.3829 \\ 
    InsantDrag~\cite{shin2024instantdrag} & 0.4810 & 163.6477 & 34.6370 \\
    Blended LDM~\cite{avrahami2023blended} & 0.5012 & 185.0291 & 41.3255 \\
    MOFA-Video~\cite{niu2024mofa} & 0.3283	& 149.1247 & 20.9583 \\
    \textbf{Ours} & \textbf{0.2397} & \textbf{132.1145} & \textbf{13.0228} \\
    \bottomrule
\end{tabular}
\vspace{\tabmargin}
\label{tab:quan.comp}
\end{table}

%% file: tables/tab_quan_ablation_data_comp.tex
\begin{table}[t]
\caption{
    Quantitative results for training with different sets of training data. 
}
\vspace{\tabmargin}
\centering
\setlength{\tabcolsep}{0.25em}
\small
\begin{tabular}{cc cc cc} 
    \toprule
    Data setting & LPIPS $\downarrow$ & FID $\downarrow$ \\
    \midrule
    Transform source & 0.3874 & 151.6589 \\
    Transform source + MF & 0.3321 & 145.3312 \\
    \textbf{Transform source + Transform Target + MF} &\textbf{0.2397} & \textbf{132.1145}  \\
    \bottomrule
\end{tabular}
\vspace{\tabmargin}
\label{tab:quan.ablation}
\end{table}

%% file: tables/tab_quan_ablation_cond.tex
\begin{table}[t]
\caption{
    \editI{\textbf{Ablation study of conditioning.} We evaluate the effect of different conditioning mechanisms with two mask configurations: ({0.0,0.5,1.0}) v.s. ({0.0,1.0}), and the impact of dropout of image features.}
}
\vspace{\tabmargin}
\centering
\setlength{\tabcolsep}{0.25em}
\small
\begin{tabular}{cc cc cc cc} 
    \toprule
    Configurations & Mask (0.0, 1.0) & Without dropout & \textbf{Ours} \\
    \midrule
    FID $\downarrow$ & 17.5615 & 21.2870 & \textbf{13.0228} \\ 
    \bottomrule
\end{tabular}
\vspace{\tabmargin}
\vspace{-3mm}
\label{tab:quan.ablation.cond}

\end{table}

%% file: 5-conclusion.tex
\section{Conclusion}
\label{sec:conclusion}
We introduced 3D-Fixup, a workflow that tackles the problem of 3D-aware image editing. To make 3D-aware image editing efficient, we adopt a feedforward method. To train such a model, data is crucial. Since collecting data with corresponding 3D edits is time-consuming and expensive, we developed an automatic framework to collect suitable data from real-world videos. The resulting method bridges the gap between 2D image editing and 3D transformations, enabling realistic edits that preserve content identity while maintaining fidelity across various perspectives. %

We demonstrated the effectiveness of our approach through extensive experiments, showcasing its ability to handle large out-of-plane rotations and translations, as well as challenging scenarios involving significant pose changes. Quantitative evaluations using LPIPS and FID metrics validate the realism and accuracy of our edits, outperforming state-of-the-art baselines such as Magic Fixup. 

\editI{We found that intricate details, such as sprinkles on donuts or textures on clothing, are sometimes not preserved well, likely due to image encoder limitations. Additionally, 3D-Fixup produces suboptimal results when $I_{\text{guide}}$ is of low quality. This occurs when the image-to-3D step performs poorly due to occlusion, incompleteness, or a suboptimally detected mask, which can be mitigated by outpainting masks/objects.}
Future work may explore extending the framework to handle more complex scenes with multiple objects and refining the underlying 3D priors to enhance generalization across diverse datasets. %
\begin{acks}
Work supported in part by NSF grants 2008387, 2045586, 2106825, MRI 1725729, NIFA award 2020-67021-32799, and the Amazon-Illinois Center on AI for Interactive Conversational Experiences.
\end{acks}

%% file: figs/3-fig-qual-ours.tex
\begin{figure*}[htbp]
    \centering
    \includegraphics[width=\linewidth]{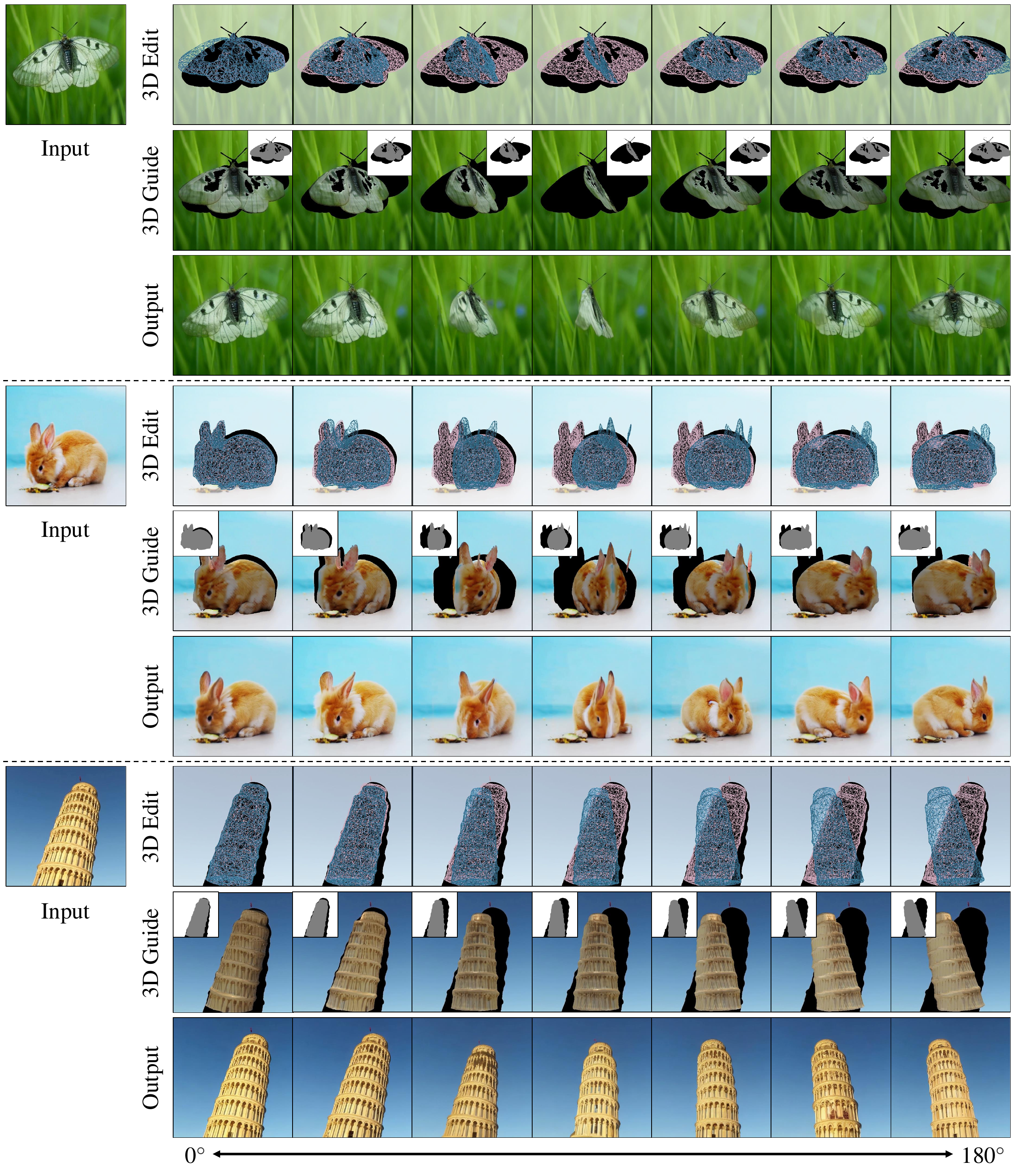}
    \vspace {-6
    mm}
    \caption{
    \textbf{3D editing with continuous rotations.} We demonstrate that the proposed method can handle extensive 3D edits on common objects. In each scenario, we progressively increase the rotation from 0 to 180 degrees along the y-axis, applying it to the reconstructed mesh to generate the 3D-transformation-based image guidance. The results show that our method can handle large out-of-plane 3D rotations during editing.
    }
    \label{fig:3.qual.ours}
\end{figure*} 

%% file: figs/3-fig-qual-comp-mf.tex
\begin{figure*}[htbp]
    \centering
    \includegraphics[width=.95\linewidth]{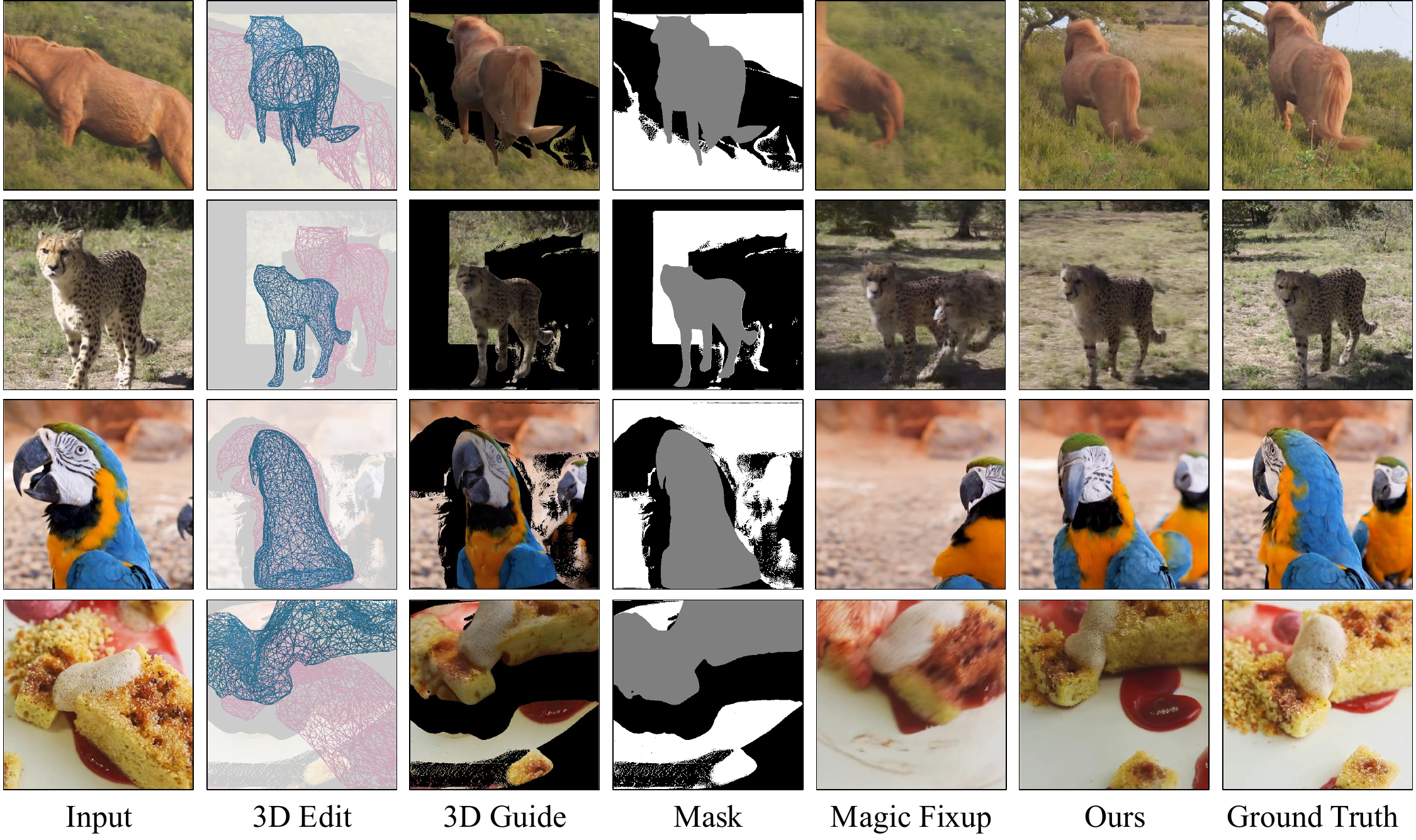}
    \caption{
        \textbf{Comparison with Magic Fixup.} The results demonstrate that the proposed method achieves more realistic outputs by leveraging the 3D-transformation-based guidance. For instance, our method effectively handles pose changes, such as adjusting the camera's viewing direction \editII{for the cakes and jaguar}, or modifying the poses of the horse \editII{and parrot}.
    }
    \vspace \figmargin
    \label{fig:3.qual.comp.mf}
\end{figure*}